\documentclass[letterpaper, 10 pt, conference]{ieeeconf}

\IEEEoverridecommandlockouts                  

\usepackage{graphicx}
\usepackage{tabularx}

\usepackage{cite}
\usepackage{amsmath,amssymb,amsfonts}
\usepackage{algorithmic}
\usepackage{textcomp}
\usepackage{pifont} %

\usepackage{balance}
\usepackage{comment}
\usepackage[ruled,vlined]{algorithm2e}
\usepackage{caption}

\usepackage{multirow}
\usepackage{array}
\usepackage{subcaption}
\usepackage{arydshln}
\usepackage{tensor}
\newcommand{\eg}{\emph{e.g.},}

\usepackage{hyperref}
\usepackage{float}
\usepackage{rotating}
\usepackage{nicematrix}  %
\usepackage{mathtools}
\usepackage{amsmath}
\usepackage{pifont}
\usepackage{cuted}
\usepackage{dblfloatfix}
\usepackage{gensymb}
\usepackage[ruled,vlined]{algorithm2e}
\usepackage{titlesec}
\usepackage{amssymb}
\usepackage{graphicx}
\let\labelindent\relax
\usepackage{enumitem}

\def\dsname{{WildCross}}

\usepackage{fancyhdr}
\fancypagestyle{withfooter}{
  \fancyhf{} 

  \fancyfoot[C]{\footnotesize Accepted to the IEEE ICRA Workshop on Open Challenges for Rigorous Robot Perception 2026}
}

\bstctlcite{IEEEexample:BSTcontrol}
\title{\LARGE \bf
Cross-Modal Benchmarking for Robotic Perception in Natural Environments
}
\author{ David Hall$^{1}$, Joshua Knights$^{2}$, Mark Cox$^{1}$, Peyman Moghadam$^{1,3}$
\thanks{$^{1}$ CSIRO Robotics, CSIRO, Australia. E-mail: {\tt\footnotesize \emph{firstname.lastname}@csiro.au}}
\thanks{$^{2}$ University of Sydney (USyd), Australia.}
\thanks{$^{3}$ Queensland University of Technology (QUT), Australia.}
}

\begin{document}

\maketitle
\thispagestyle{withfooter}
\pagestyle{withfooter}

\begin{abstract}
Natural environments present a complex challenge to robotics perception systems.
Current models, particularly vision foundation models, are largely trained on structured, urban environments leading to weaknesses in their perception for field robotics tasks.
We showcase the limitations of current models using our recently released \textit{WildCross} benchmark,  a new cross-modal benchmark for place recognition and metric depth estimation in large-scale natural environments.
{WildCross} comprises over 476K sequential RGB frames with semi-dense depth and surface normal annotations, each aligned with accurate 6DoF pose and synchronized dense lidar submaps.
In this work, we provide an expanded analysis of the benchmark results from the recent WildCross benchmark, with particular emphasis on expanded metric depth estimation experiments.
Access to the code repository and dataset for this work can be found at \href{https://csiro-robotics.github.io/WildCross}{https://csiro-robotics.github.io/WildCross}.
\end{abstract}
\section{INTRODUCTION}
\label{sec:intro}
Autonomous robots are increasingly deployed in unstructured and natural environments for applications such as agriculture, environmental monitoring, and search and rescue~\cite{oliveira2021advancesagri}. 
However, progress in robotic navigation and perception tasks remains heavily dependent on public datasets, given the high cost and logistical challenges of large-scale field trials. Benchmarks such as KITTI~\cite{geiger2013vision} and Oxford RobotCar~\cite{maddern20171} have been instrumental in advancing the field, but they are predominantly captured in structured urban or indoor settings~\cite{silberman2012indoor,geiger2013vision,maddern20171}. In contrast, natural environments are characterized by irregular terrain, dense vegetation, narrow trails, and complex occlusions, rendering existing datasets insufficient for evaluating robotic autonomy in environments where it is most urgently required.  Concurrently, the robotics and computer vision communities are placing increasing emphasis on bridging 2D and 3D scene understanding, exemplified by recent advances in learning-based 3D reconstruction~\cite{wang2025vggt} and cross-modal place recognition~\cite{shubodh2024lip}. To support these developments, datasets must provide accurate ground truth across both 2D and 3D modalities under the added complexity of natural scenes. 
Such datasets enable training of new high-quality models for field robotics and allow for analysis on the current gaps that exist between current state-of-the-art systems and the needs of robotic perception in natural environments.

This paper presents a companion to our original \textit{WildCross} paper~\cite{wildcross} where we provide a condensed overview of the WildCross data with an expanded analysis of some of the original paper's benchmark results.
These showcase how WildCross moves beyond our past Wild-Places~\cite{knights2022wild} dataset, providing a benchmark for visual and cross-modal place recognition and monocular depth estimation experiments within natural environments. 
In particular, we expand the original paper's analysis of monocular depth estimation, examining the new DepthAnythingV3's monocular metric depth estimation model and fine-tuning DepthAnythingV2 \cite{yang2024depth} models using pseudo-ground-truth (PGT) estimates that fuse monocular depth estimation model outputs with WildCross' semi-dense ground-truth (GT) depth data.

\begin{figure}
    \centering
    \includegraphics[width=0.6\linewidth]{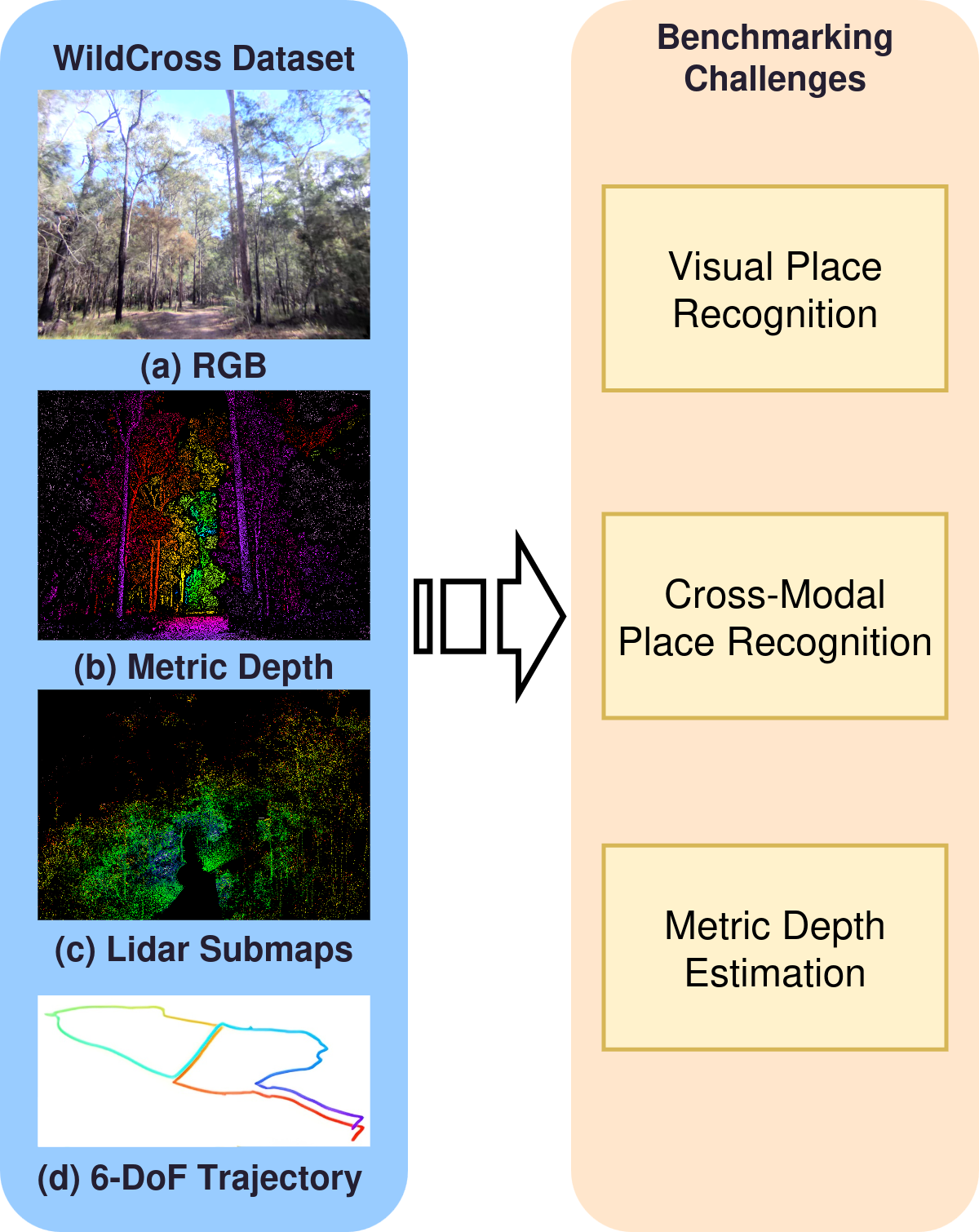}
    \caption{Overview of benchmarking using WildCross data.  WildCross contains RGB images with corresponding sparse metric depth measurements, lidar submaps and 6-DoF poses for eight traversals of bushwalking trails near Brisbane, Australia, allowing for the benchmarking of a number of critical robotics perception tasks in complex natural environments.}
    \vspace{-5mm}
    \label{fig:hero}
\end{figure}
\section{WILDCROSS}
\label{sec:dataset}

The WildCross benchmark \cite{wildcross} leverages the raw data from the Wild-Places~\cite{knights2022wild} LPR dataset and extends it into a cross-modal benchmark for place recognition and metric depth estimation through two main advances, complementary to WildScenes~\cite{vidanapathirana2025wildscenes}, which focuses on 2D and 3D semantic segmentation in the same natural environments.
Firstly, original traversals are reprocessed to produce sequential RGB frames at 15Hz, with accurate 6DoF ground truth poses synchronized with dense 3D lidar submaps in the same environment.  
Secondly, an annotation pipeline generates semi-dense metric depth and surface normal image mappings for every RGB frame. 
This enables WildCross to be used as a benchmark for evaluating performance on the tasks of visual and cross-modal place recognition, in addition to metric depth estimation in complex unstructured natural environments.

For consistency with Wild-Places~\cite{knights2022wild} and WildScenes~\cite{vidanapathirana2025wildscenes}, WildCross \cite{wildcross} adopts the notation V-XX and K-XX to denote sequence XX on data captured at the Venman and Karawatha locations respectively.  The traversals follow a consistent pattern: in both environments, Sequence 02 corresponds to the reverse trajectory of Sequence 01, Sequence 03 follows an alternate extended route, and Sequence 04 repeats the route of Sequence 01. 
The reverse trajectory (Sequence 02) follows approximately the same path as Sequence 01 but in the opposite direction. Since only a forward-facing camera is used, images taken at the same location across the two sequences show little visual overlap. 

Color images are obtained at 15Hz from a forward-facing camera on the sensor payload, and are rectified using distortion parameters after sensor calibration.  Lidar submaps are sampled along the trajectory of a global accumulated point cloud map generated using lidar-inertial SLAM~\cite{ramezani2022wildcat}, following the approach used in Wild-Places~\cite{knights2022wild}.  For each camera frame, WildCross also provides semi-dense metric depth and surface normal annotations which can be beneficial for domains such as monocular depth estimation and neural fields.  
These are generated from the full global point-map with considerations made for visibility to remove occluded points from each annotated frame.  
For full details on this process and further statistics on \dsname{}, we refer readers to the original paper~\cite{wildcross}.
\section{EXPERIMENTS}

\begin{table}[t!]
    \centering
    \addtolength{\tabcolsep}{-0.4em}
    \begin{NiceTabular}{c|cccc|cc|cc}
         \Block{2-1}{Split} &\Block{2-1}{ V/K-01} & \Block{2-1}{V/K-02} & \Block{2-1}{V/K-03} & \Block{2-1}{V/K-04} & \Block{1-2}{Lidar} && \Block{1-2}{Camera} &   \\
         & &&&& Train&Test&Train&Test \\\hline 
         01 & \textbf{Test} & Train & Train & Train &49.8K & 13.5K& 374.6K & 101.4K \\
         02 & Train & \textbf{Test} & Train & Train &48.7K & 14.6K& 366.3K & 109.7K \\
         03 & Train & Train & \textbf{Test} & Train &39.7K & 23.6K& 298.0K & 177.9K \\
         04 & Train & Train & Train & \textbf{Test} &51.8K & 11.5K& 389.1K & 86.9K \\
         
    \end{NiceTabular}
    \caption{Train/test cross-fold splits for \dsname{}.}
    \label{tab:splits}
    \vspace{-7mm}
\end{table}

\label{sec:experiments}

\subsection{Training and Testing Splits}
Unlike the training splits used for LPR in Wild-Places~\cite{knights2022wild}, \dsname{} utilises a cross-fold training and evaluation setup for benchmarking VPR and CMPR networks.
In this setup, training and evaluation follow a four-fold cross-split design. In each split, the sequences with the same index (\eg{} Split-1, V-01, and K-01) from both environments are held out together for evaluation, while the remaining sequences are used for training. 
Table \ref{tab:splits} reports the number of training and testing samples in each split. During evaluation, the held-out sequences are used for intra-sequence place recognition and also serve as queries for inter-sequence recognition, with the training sequences acting as the database.

\subsection{Visual Place Recognition (VPR)}
\label{subsec:vpr_exp}

We evaluate four state-of-the-art methods for VPR: 
NetVLAD \cite{arandjelovic2016netvlad}, MixVPR~\cite{ali2023mixvpr}, DINOv2-SALAD (SALAD)~\cite{Izquierdo_CVPR_2024_SALAD}, and Bag-of-Queries (BoQ)~\cite{ali2024boq}.  
Unlike LPR, positive training pairs for VPR cannot be formed solely using a distance threshold. The limited field-of-view of the camera and the presence of reverse revisits within and across sequences can result in false positives, where two images selected as a pair share little or no visual overlap. To mitigate this, and inspired by ~\cite{Berton_CVPR_2022_CosPlace}, we define positive training pairs in \dsname{} as images whose camera poses are within 5m distance and 15\degree~bearing of each other, and negative pairs are those separated by more than 50m. At evaluation, a retrieved image is considered a correct match if its pose lies within 25m of the query. We report results under two evaluation settings: zero-shot and fine-tuned. In the \textbf{zero-shot} setting, each method is evaluated on \dsname{} using its best released pretrained model, without any additional training on \dsname{}.  In the \textbf{fine-tuned} setting, the same methods are further fine-tuned on the \dsname{} training splits before evaluation. This measures their in-domain performance once exposed to data from natural environments. Reporting both settings highlights the gap between cross-domain generalization (urban-to-natural) and in-domain adaptation, providing a comprehensive benchmark of VPR in natural environments.\looseness=-1

\begin{table*}[t]
    \centering
    \addtolength{\tabcolsep}{-0.25em}
    \resizebox{\textwidth}{!}{
    \begin{NiceTabular}{ll | cccccccccccccccc|cc}
         \Block{2-2}{Method} && 
         \Block{1-2}{V-01} && \Block{1-2}{V-02} && \Block{1-2}{V-03} && \Block{1-2}{V-04} &&
         \Block{1-2}{K-01} && \Block{1-2}{K-02} && \Block{1-2}{K-03} &&\Block{1-2}{K-04} &&\Block{1-2}{Average}\\
         && R1 & R5 & R1 & R5 & R1 & R5 &R1 & R5 
         & R1 & R5 & R1 & R5 & R1 & R5 & R1 & R5 & R1 & R5 \\
        \hline 
        \Block{4-1}<\rotate>{Zero-shot}
        &NetVLAD~\cite{arandjelovic2016netvlad} &47.24&51.91&\textbf{52.44}&\textbf{64.70}&7.040&14.65&38.98&47.91&47.09&59.10&55.70&73.86&13.76&23.34&51.98&55.67&39.28&48.89\\
        &MixVPR~\cite{ali2023mixvpr}            &51.86&57.39&44.60&50.63&7.33&13.80&44.60&50.84&74.89&81.88&\textbf{69.53}&\textbf{85.18}&24.88&32.34&\textbf{57.12}&\textbf{59.58}&\textbf{46.85}&\textbf{53.96}\\
        &SALAD~\cite{Izquierdo_CVPR_2024_SALAD} &50.64&57.01&44.60&49.06&\textbf{13.05}&\textbf{19.58}&43.72&48.53&78.48&83.50&54.71&68.16&\textbf{26.67}&\textbf{32.78}&54.93&59.31&45.85&52.24\\
        &BoQ~\cite{ali2024boq}                  &\textbf{54.65}&\textbf{62.48}&48.22&55.88&11.62&17.69&\textbf{46.66}&\textbf{52.47}&\textbf{80.99}&\textbf{84.22}&45.06&55.62&26.61&31.36&55.41&58.19&46.15&52.24\\
        \hline 
        \Block{4-1}<\rotate>{Fine-tuned}
        &NetVLAD~\cite{arandjelovic2016netvlad} &68.95&72.16&69.89&74.05&18.09&22.72&59.90&66.02&83.86&87.53&86.32&91.57&26.82&32.35&56.32&59.96&58.77&63.30\\
        &MixVPR~\cite{ali2023mixvpr}            &66.16&68.52&68.86&72.78&17.80&25.01&52.90&57.34&84.66&87.89&87.99&89.59&30.40&38.22&60.87&\textbf{65.04}&58.71&63.05\\
        &SALAD~\cite{Izquierdo_CVPR_2024_SALAD} &69.04&72.68&72.24&\textbf{80.33}&23.30&29.48&58.96&64.65&86.37&89.33&88.98&91.26&\textbf{34.05}&\textbf{38.83}&\textbf{61.13}&62.53&61.76&66.14\\
        &BoQ~\cite{ali2024boq}                  &\textbf{70.22}&\textbf{74.61}&\textbf{72.84}&77.67&\textbf{28.68}&\textbf{38.87}&\textbf{62.40}&\textbf{68.71}&\textbf{87.62}&\textbf{89.51}&\textbf{90.81}&\textbf{92.93}&32.26&38.05&60.49&62.90&\textbf{63.17}&\textbf{67.91}\\
    \end{NiceTabular}
    }
    \caption{Intra-sequence VPR results on \dsname{} for zero-shot and fine-tuned networks.}
    \label{tab:vpr_intra}
    \vspace{-6mm}
\end{table*}

\begin{table}[t]
    \centering
    \addtolength{\tabcolsep}{-0.35em}
    \begin{NiceTabular}{ll|cccc|cc}
         \Block{2-2}{Method (Backbone)} && \Block{1-2}{Venman} && \Block{1-2}{Karawatha} && \Block{1-2}{Average}  \\
         && R1 & R5 & R1 & R5 & R1 & R5 \\
         \hline 
         \Block{4-1}<\rotate>{Zero-Shot} 
         &NetVLAD~\cite{arandjelovic2016netvlad}    &25.86&43.94&16.15&29.73&21.00&36.84\\
         &MixVPR~\cite{ali2023mixvpr}               &54.10&61.41&35.73&44.31&44.92&52.86 \\
         &SALAD~\cite{Izquierdo_CVPR_2024_SALAD}    &57.49&64.49&41.27&50.14&49.38&57.32\\
         &BoQ~\cite{ali2024boq}                     &\textbf{61.62}&\textbf{67.98}&\textbf{45.89}&\textbf{54.98}&\textbf{53.76}&\textbf{61.48} \\
         \hline
         \Block{4-1}<\rotate>{Fine-tuned}
         &NetVLAD~\cite{arandjelovic2016netvlad}    &64.31&67.49&46.94&52.43&55.63&59.96\\
         &MixVPR~\cite{ali2023mixvpr}               &65.30&68.58&50.24&55.80&57.77&62.19\\
         &SALAD~\cite{Izquierdo_CVPR_2024_SALAD}    &68.54&71.86&54.29&59.86&61.41&65.86\\
         &BoQ~\cite{ali2024boq}                     &\textbf{68.66}&\textbf{72.01}&\textbf{55.07}&\textbf{60.37}&\textbf{61.87}&\textbf{66.19}\\
    \end{NiceTabular}
    \caption{Inter-Sequence VPR Results on \dsname{} for zero-shot and fine-tuned networks. }
    \label{tab:vpr_inter}
    \vspace{-4mm}
\end{table}

\subsection{Cross-Modal Place Recognition (CMPR)}

Cross-modal place recognition (CMPR) aims to localize across different sensing modalities, such as retrieving lidar submaps given visual queries. This task is particularly challenging in natural environments, where structural complexity and viewpoint variation exacerbate the difficulty of aligning cross-modal features. To explore this task on \dsname{}, we evaluate a lightly modified version of LIP-Loc~\cite{shubodh2024lip}. Within our cross-fold training and testing regime, we evaluate the inter-sequence CMPR performance using the images from the unseen sequences as queries and lidar submaps from all sequences in an environment as databases.  
We refer readers to the original \dsname{} paper~\cite{wildcross} for full details on how LIP-Loc was modified for our experiments.
Of relevance to this paper, we evaluate LIP-Loc using three different pre-trained backbone encoders, namely ResNet50~\cite{he2016deep}, DINOv2~\cite{oquab2023dinov2}, and DINOv3~\cite{simeoni2025dinov3}.

\subsection{Metric Depth Estimation}
\dsname{} also supports research in metric depth estimation. 
We evaluate this task using DepthAnythingV2~\cite{yang2024depth} as a representative state-of-the-art baseline. 
For this experiment, sequences V-01 and K-01 are held out for testing, K-02 is used for validation, and the remaining sequences are used for training.
We report three common metrics:
threshold accuracy ($\delta_1$), which measures the percentage of the predicted pixels whose depth differs from ground truth by no more than 25\%, Absolute Relative Error (AbsRel), which quantifies the average relative difference between predicted and true depths, and Root Mean Square Error (RMSE), which measures the overall deviation of predictions from the ground truth.
Metrics are only calculated over pixels with known GT values.

As with VPR, we report results under both zero-shot and fine-tuned settings. In the zero-shot setting, we directly evaluate the released model trained on KITTI~\cite{geiger2013vision} and VirtualKitti~\cite{gaidon2016virtual}, thereby assessing Out-Of-Domain (OOD) generalization from urban to natural environments. 
For fine-tuning we use a handful of different methods.
The baseline fine-tuning is that reported in the original \dsname{} paper with the model fine-tuned using the semi-dense GT depth data provided by \dsname{}.
However, past works have highlighted that training on anything but fully-dense data can cause a degradation in a model's ability to estimate fine-grained depth details (e.g. small leaves)\cite{yang2024depth,lin2025depth}.
To this end, we explore training using pseudo-GT (PGT) depth images produced by rescaling outputs from pre-trained depth estimation models.
The first PGT used is the output from the new DepthAnything metric model \cite{lin2025depth} which is designed to produce accurate depth estimates in metres when provided camera focal lengths.
As focal lengths are provided by WildCross, we investigate both the zero-shot accuracy of the DAv3 model output, and how effective the outputs can be for fine-tuning other depth estimation models.
The second PGT is to align monocular depth estimation model outputs with the \dsname{} using RANSC least squares as done in \cite{lin2025depth}.
The final PGT is produced using the Prior Depth Anything (PriorDA)~\cite{wang2025depth} default network which fuses outputs from a DAv2 model with any level of sparse GT depth to provide dense and well-scaled data.

\section{RESULTS}
\label{sec:results}

\subsection{Visual Place Recognition}
\label{subsec:results/vpr}
Tables~\ref{tab:vpr_intra} and~\ref{tab:vpr_inter} summarize intra- and inter-sequence VPR performance on \dsname{}. Fine-tuning consistently improves performance across all methods, showcasing the value of training on in-domain natural environment data. Nevertheless, even the strongest overall method, BoQ~\cite{ali2024boq}, achieves only 63.17\% R1 for intra-sequence and 61.87\% for inter-sequence evaluation. In comparison, on established urban benchmarks such as Pittsburgh~\cite{torii2013visual} and MSLS~\cite{warburg2020mapillary}, the same method exceeds 90\% R1.

One of the primary challenges the \dsname{} benchmark provides is from the prevalence of reverse revisits in both intra- and inter-sequence evaluation.  In intra-sequence place recognition, we note the lowest performance occurs on sequences with the largest number of reverse revisits (V-03 and K-03); in addition, as is elaborated on in greater detail in the main WildCross~\cite{wildcross} paper, inter-sequence performance drops remarkably when query and database are from from reverse trajectory sequences (e.g. V-02 queries for V-01 database) where images of the same location have little to no overlap in FoV.
These results highlight that unstructured natural environments, particularly reverse revisits, remain a persistent challenge for state-of-the-art VPR methods.

\subsection{Cross-Modal Place Recognition}
\label{subsec:results/cmpr}

\begin{table}[t]
    \centering
    \addtolength{\tabcolsep}{-0.35em}
     \begin{NiceTabular}{l|cccccc}
         \Block{2-1}{Method (Backbone)} & \Block{1-2}{Venman} && \Block{1-2}{Karawatha} && \Block{1-2}{Average}  \\
         & R1 & R5 & R1 & R5 & R1 & R5 \\
         \hline 
         LIP-Loc (ResNet50) &40.16&54.45&34.25&48.91&37.20&51.68 \\
         LIP-Loc* (DINOv2-s) &52.55&62.71&45.26&\textbf{57.40}&48.90&60.06\\
         LIP-Loc* (DINOv3-s) &\textbf{56.54}&\textbf{63.19}&\textbf{48.16}&57.06&\textbf{52.35}&\textbf{60.12}\\
    \end{NiceTabular}
    \caption{CMPR Results on \dsname{}.  * ViT-S for the pretrained model backbone.}
    \label{tab:cmpr_results}
    \vspace{-2mm}
\end{table}

Table~\ref{tab:cmpr_results} reports CMPR results on \dsname{}. Performance across all configurations remains limited, with the best result obtained by LIP-Loc using DINOv3 pretraining, which achieves an average R1 score of 52.35\%.  We predict that significant future progress in this task will likely require approaches which explicitly address the domain gap between 2D image features and 3D structural representations, rather than relying on backbone architectures or purely vision-based large-scale pre-training alone. 

\subsection{Metric Depth Estimation}
\begin{figure*}
    \centering
    \includegraphics[width=0.8\linewidth]{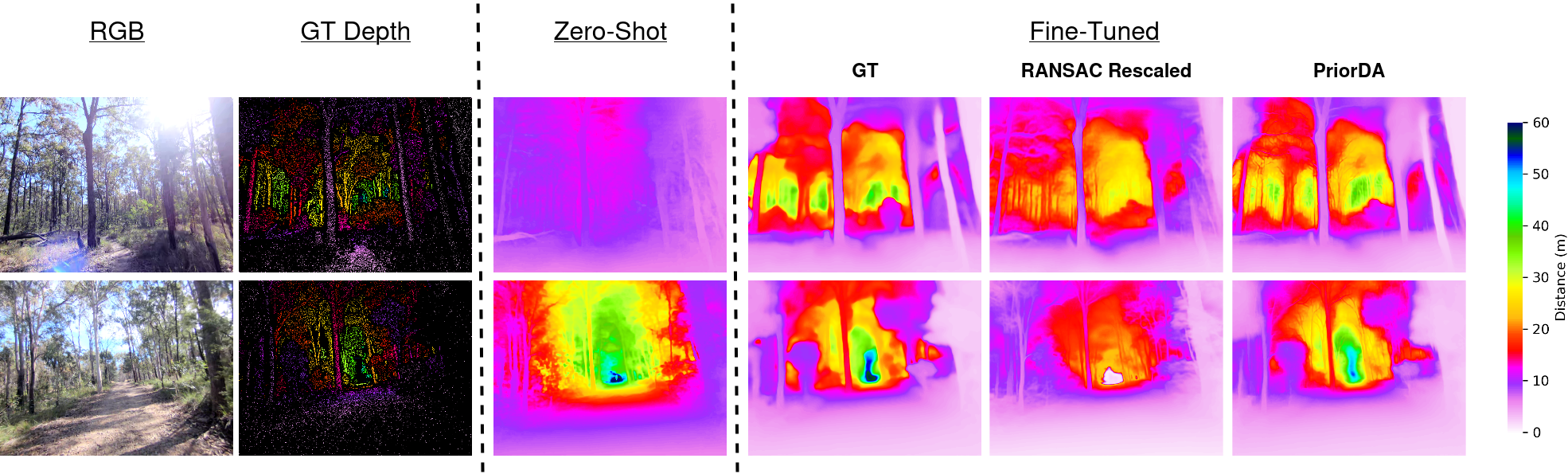}
    \caption{Example outputs for DA2-ViT-S model compared to ground-truth depth data. Left shows RGB and GT depth. Centre shows output of zero-shot pre-trained DA2-ViT-S model. Right shows DA2-ViT-S model fine-tuned on GT, RANSAC-rescaled PGT, and PriorDA PGT data respectively. Best viewed digitally.}
    \vspace{-7mm}
    \label{fig:depthvis}
\end{figure*}
\begin{table}[t]
    \centering
    \begin{NiceTabular}{ll|ccc}
         & Method (Backbone) & $\delta_1\uparrow$ & AbsRel$\downarrow$ & RMSE$\downarrow$\\
         \hline 
         \Block{3-1}{Zero-Shot} 
         & DA2 (ViT-S) &\textbf{0.284}&\textbf{0.558}&\textbf{7.651} \\
         & DA2 (ViT-B) &0.222&0.769&7.915 \\
         & DA2 (ViT-L) &0.074&1.478&13.734\\
         \hline 
         \Block{3-1}{Fine-Tuned} 
         &  DA2 (ViT-S) &0.746&0.172&3.412 \\
         &  DA2 (ViT-B) &0.766&0.167&3.289 \\
         &  DA2 (ViT-L) &\textbf{0.789}&\textbf{0.157}&\textbf{3.150} \\
    \end{NiceTabular}
    \caption{Depth estimation zero-shot and fine-tuned results}
    \label{tab:depthpred}
    \vspace{-2mm}
\end{table}

\begin{table}[t]
    \centering
    \begin{NiceTabular}{l|ccc}
         Fine-tuning Data & $\delta_1\uparrow$ & AbsRel$\downarrow$ & RMSE$\downarrow$\\
         \hline 
         None &0.284&0.558&7.651 \\
         \dsname{} - GT &0.746&0.172&3.412 \\
         RANSAC-DA2-ViT-S &0.457&0.350&6.595\\
         RANSAC-DA2-ViT-L &0.505&0.318& 5.729 \\
         RANSAC-DA3-ViT-L-Metric & 0.556 & 0.316 & 5.492 \\
         DA3-ViT-L-Metric & 0.413 & 0.423 & 5.874 \\
         PriorDA-ViT-B & 0.728 & 0.175 & 3.521 \\
    \end{NiceTabular}
    \caption{Results for fine-tuning DA2-ViT-S models~\cite{yang2024depth} on different GT and Pseudo-GT data.}
    \label{tab:expanded_depth_pred}
    \vspace{-2mm}
\end{table}

\begin{table}[t]
    \centering
    \begin{NiceTabular}{l|ccc}
         Pseudo GT & $\delta_1\uparrow$ & AbsRel$\downarrow$ & RMSE$\downarrow$\\
         \hline 
         DA3-ViT-L-Metric & 0.414 & 0.380 & 6.086 \\
         RANSAC-DA2-ViT-S &0.522&0.364&6.764\\
         RANSAC-DA2-ViT-L &0.573&0.334&5.883 \\
         RANSAC-DA3-ViT-L-Metric & 0.601 & 0.354 & 6.184 \\
         PriorDA-ViT-B & \textbf{0.980} & \textbf{0.044} & \textbf{1.156} \\
    \end{NiceTabular}
    \caption{Evaluation of Pseudo-GT depths against true GT}
    \label{tab:pseudo-gt}
    \vspace{-6mm}
\end{table}
Table~\ref{tab:depthpred} reports zero-shot and fine-tuned metric depth prediction results on \dsname{}. Fine-tuning with our depth annotations consistently improves the performance of DepthAnythingV2~\cite{yang2024depth} across all backbones, with larger ViT backbones yielding stronger results. 
However, qualitative results in Figure~\ref{fig:depthvis} show that fine-tuning on sparse data improves the overall scale of the predictions but reduces fine-grained detail (e.g. leaf distinction) compared to the pretrained model. 
This trait has been highlighted in previous depth estimation works~\cite{yang2024depth,lin2025depth}, motivating research into using depth estimates to create PGT data to train on.

In Table~\ref{tab:pseudo-gt}, we evaluate how close to the original GT the PGT approaches tested were.
Here we see that the new DepthAnythingV3 metric model, with additional camera calibration information, is able to exceed zero-shot DepthAnythingV2 models in Table~\ref{tab:depthpred} while not exceeding results of fine-tuned ones.
Furthermore, we see that RANSAC-rescaled outputs consistently align better to the GT data than the original zero-shot outputs shown in Table~\ref{tab:depthpred} with larger/more recent models doing better and DA3-ViTL-Metric getting the highest results of the RANSAC-rescaled data.
However, it is clear that PriorDA is best at utilising the underlying GT data, achieving strong alignment with for pixels with known depth values.

Table~\ref{tab:expanded_depth_pred} and Figure~\ref{fig:depthvis} show the impact from fine-tuning a model using PGT data.
Fine-tuning using PGT data, while qualitatively retaining sharpness of depth image outputs, never exceeds the raw metric accuracy that can be achieved using GT data for fine-tuning.
However, of the methods tested, PriorDA PGT appears to provide the best balance between the methods, still attaining sharp segmentation details, while not suffering a large drop in quantitative evaluation metrics.
While these are promising initial results, they also show that WildCross provides a challenging benchmark for metric depth estimation.
We attribute this to the domain as forests contain large disparities in depth between trees in the scene with very few visual queues to clearly indicate them.
Monocular depth estimation is therefore not yet sufficiently reliable within this setting to replace traditional sensors for robotics applications.
\section{CONCLUSION}
This paper provides an extended analysis of benchmark methods for VPR, CMPR, and depth estimation using the new WildCross benchmark. 
Our tests show how WildCross presents a unique challenge for VPR and CMPR systems due to an under-examined domain and reverse revisits.
We expand our investigation into depth estimation models, examining the usefulness of Pseudo-GT depth estimation training data.
This showed how current state-of-the-art depth estimation models do not cope with forest structures which contain dramatic depth differences between trees without clear visual cues for the model to work from.
In this domain, monocular depth estimation still remains insufficient for replacement of other robotic sensors that measure depth such as LiDAR.
Finally, we found that fine-tuning using PriorDA pseudo-GT images achieves the best trade-off for producing the most accurate depth estimates while maintaining fine-detail visual features over just training on measured GT data.
The results presented provide a starting benchmark for future work in this unique and challenging domain.

\bibliographystyle{IEEEtran}
\bibliography{ref}

\end{document}